\documentclass[journal]{IEEEtran}
\usepackage{cite}
\usepackage{amsmath,amssymb,amsfonts}
\usepackage{algorithmic}
\usepackage{graphicx}
\usepackage{textcomp}
\usepackage{multirow}
\usepackage{booktabs}
\usepackage{subcaption} 
\usepackage{xcolor}

\newcommand{\added}[1]{\textcolor{black}{#1}}

\newcommand{\rev}[1]{}

\usepackage{dsfont}
\usepackage[normalem]{ulem}
\usepackage[table]{xcolor}
\definecolor{LightGray}{gray}{0.95}

\usepackage{hyperref}

\def\BibTeX{{\rm B\kern-.05em{\sc i\kern-.025em b}\kern-.08em
    T\kern-.1667em\lower.7ex\hbox{E}\kern-.125emX}}
\begin{document}

\title{A 3D Cross-modal Keypoint Descriptor for MR-US Matching and Registration}
\bstctlcite{IEEEexample:BSTcontrol}
\author{Daniil Morozov, Reuben Dorent, and Nazim Haouchine
\thanks{Daniil Morozov and Nazim Haouchine are with Harvard Medical School and Brigham and Women's Hospital, Boston, MA, USA.}
\thanks{Reuben Dorent is with Inria and Sorbonne Université, Institut du Cerveau - Paris Brain Institute - ICM, CNRS, Inserm, AP-HP, Hôpital de la Pitié Salpêtrière, F-75013, Paris, France}
\thanks{Daniil Morozov is also with Technical University of Munich, Germany}
}

\maketitle

\begin{abstract}
Intraoperative registration of real-time ultrasound (iUS) to preoperative Magnetic Resonance Imaging (MRI) remains an unsolved problem due to severe modality-specific differences in appearance, resolution, and field-of-view. To address this, we propose a novel 3D cross-modal keypoint descriptor for MRI–iUS matching and registration. 
Our approach employs a patient-specific matching-by-synthesis approach, generating synthetic iUS volumes from preoperative MRI. This enables supervised contrastive training to learn a shared descriptor space. 
A probabilistic keypoint detection strategy is then employed to identify anatomically salient and modality-consistent locations. During training, a curriculum-based triplet loss with dynamic hard negative mining is used to learn descriptors that are i) robust to iUS artifacts such as speckle noise and limited coverage, and ii) rotation-invariant. At inference, the method detects keypoints in MR and real iUS images and identifies sparse matches, which are then used to perform rigid registration. Our approach is evaluated using 3D MRI-iUS pairs from the ReMIND dataset. Experiments show that our approach outperforms state-of-the-art keypoint matching methods across 11 patients, with an average precision of $69.8\%$.
For image registration, our method achieves a competitive mean Target Registration Error of 2.39 mm on the ReMIND2Reg benchmark.

Compared to existing iUS–MR registration approaches, our framework is interpretable, requires no manual initialization, and shows robustness to iUS field-of-view variation. 
\added{Code, data and model weights are available at \url{https://github.com/morozovdd/CrossKEY}.}\rev{R2}

\end{abstract}


\section{Introduction}
\label{sec:introduction}
\IEEEPARstart{M}{atching} images across different modalities remains a fundamental challenge in medical imaging due to significant appearance differences between modalities~\cite{jiang2021review}. This challenge underpins a variety of clinical and research applications, including content-based image retrieval~\cite{jiang2021review,kumar2013content}, slice-to-volume reconstruction~\cite{ferrante2017slice}, and deformable image registration~\cite{evan2021keymorph,Machado,JIE,heinrich2012mind,joutard2022driving}.

In the context of image-guided surgery, registering images from complementary modalities enables the fusion of distinct anatomical and functional information, improving the intraoperative identification of critical structures and ultimately surgical outcomes~\cite{assis2026systematic}. A representative example is neurosurgery, where real-time intraoperative ultrasound (iUS) is commonly registered with preoperative Magnetic Resonance Imaging (MRI) to compensate for intraoperative brain shift and refine tumor localization~\cite{Gonzalez-Darder2019,geshvadi2025optimizing}. 

\begin{figure}[t]
    \centering
    \includegraphics[width=1\linewidth]{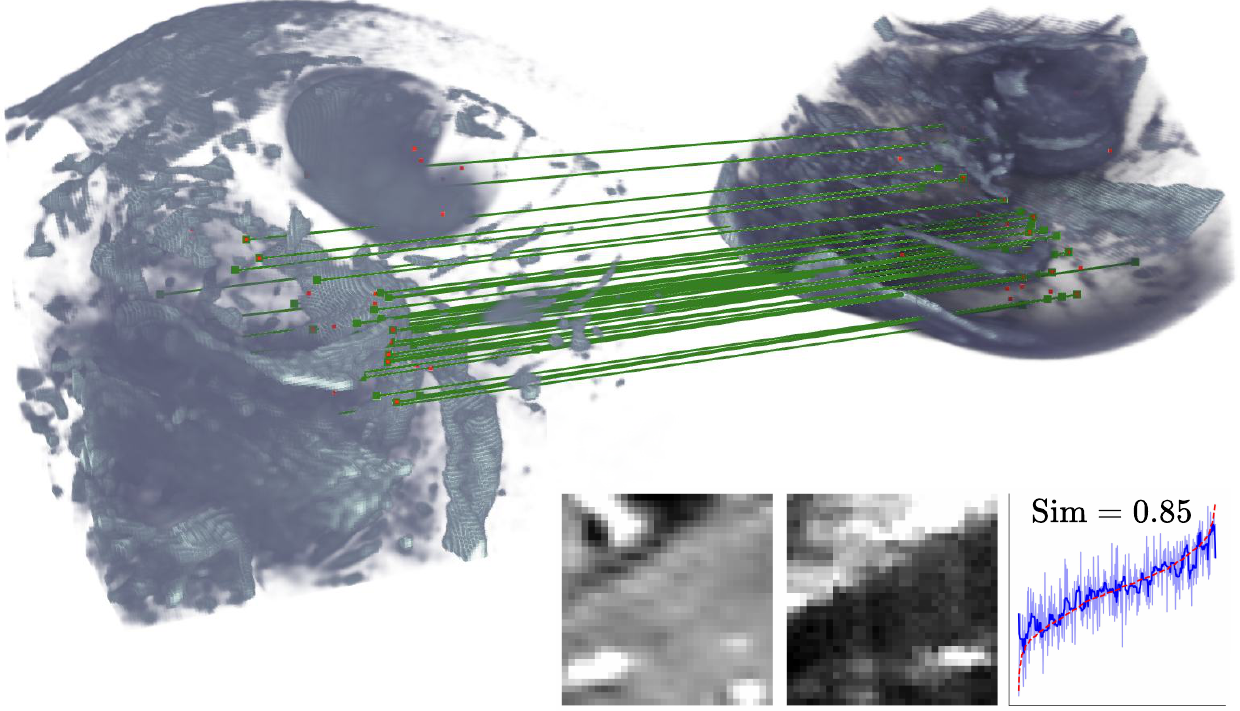}
    \caption{3D cross-modal keypoint matching between MR and iUS volumes. Bottom: Matched local patches around a keypoint pair from MR and iUS images with corresponding descriptor curves showing strong similarity agreement.}
    \label{fig:teaser}
\end{figure}
One strategy to bridge modality-specific appearance gaps is to abstract each image using keypoints~\cite{rasheed2024learning, ren2025minima, evan2021keymorph, jiang2021review}, 
 segmentations~\cite{demir2024multimodal}, or latent representations~\cite{dey2024learning,dorentUnifiedCrossModalImage2024}. When appropriately leveraged, such abstractions can facilitate multimodal medical image registration.
In this work, we focus on keypoint-based multimodal approaches, where correspondences are automatically established between a sparse set of salient keypoints detected from each image. 
Keypoints are particularly well-suited for scenarios involving partially observed or resected anatomy, such as in intraoperative imaging, and remain effective under non-rigid deformations~\cite{tursynbek2025guiding,
grewal2023automatic}. They also offer a key interpretability advantage, as matched keypoints can be directly visualized and assessed for anatomical plausibility~\cite{tursynbek2025guiding,evan2021keymorph}.
The field of multimodal keypoint matching has seen substantial investigation~\cite{jiang2021review}, with successful applications to medical imaging. Nonetheless, existing methods are often limited to 2D applications or to images where the intensity distributions across modalities remain relatively consistent. In contrast, aligning preoperative MRI with iUS images presents a particularly challenging scenario due to the large modality gap~\cite{wu2018multimodal}. MRI and iUS differ not only in terms of information they capture (morphological versus echo-based), but also in resolution, acquisition, and noise characteristics~\cite{Machado}. While MRI produces high-resolution 3D volumes with strong soft-tissue contrast derived from pulse sequence parameters, iUS provides partial and noisy views formed through acoustic wave reflections, often with limited field-of-view (FoV).
Bridging this gap requires not only handling large appearance gaps but also addressing 3D-specific challenges, including rotational and FoV variability. 

\vspace{5pt}
\noindent\textbf{Contributions.} We propose a novel cross-modal 3D keypoint descriptor specifically designed for matching preoperative MR and iUS volumes. Our main contributions are as follows:
\begin{itemize}
\item A matching-by-synthesis strategy in which synthetic iUS images are generated from the patient’s own MR volume and used to train a cross-modality descriptor network.
\item A cross-modal keypoint detector in the form of saliency heatmaps, constructed by accumulating keypoint presence across synthetic iUS and MR volumes, followed by a probabilistic aggregation to estimate keypoint saliency and consistency across modalities.
\item A supervised contrastive framework with curriculum learning, enforcing robustness to iUS appearance variability, speckle noise, rotation, and FoV changes.
\end{itemize}
This work is a substantial extension of our conference paper~\cite{rasheed2024learning}. Key improvements include: (1) an extension to fully 3D matching; (2) a novel contrastive learning formulation; and (3) expanded experimental validation, including ablation studies and quantitative evaluation on image registration. 
\added{While individual components build on prior work, the primary novelty lies in a patient-specific, 3D matching-by-synthesis framework, in which the expected appearance of an intraoperative modality is synthesized from preoperative imaging to construct a cross-modal descriptor tailored to a single patient.}\rev{R2}

 \begin{figure*}[h!]
    \centering
    \includegraphics[trim=0 8 0 0, clip, width=1\linewidth]{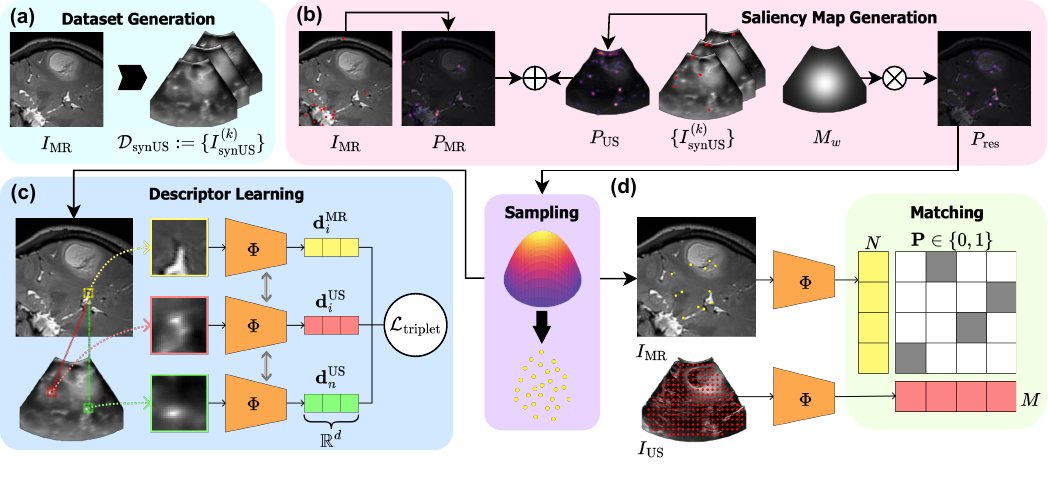}
    \caption{\textbf{Method overview.} \textit{(a)} Synthetic iUS volumes are generated from preoperative MRI using MMHVAE. \textit{(b)} A cross-modal saliency map $P_{\text{res}}$ is constructed by aggregating keypoint statistics from synthetic iUS and MRI, then modulated by a spatial prior $M_w$. \textit{(c)} A Siamese network is trained with triplet loss on multi-modal patch pairs to produce cross-modal descriptors. \textit{(d)} Descriptor matching is performed using nearest-neighbor search, followed by a partial assignment between sampled keypoints in MRI and iUS. Keypoints are sampled from the learned saliency distribution in MRI and uniformly in the real iUS.}

    \label{fig:overview}
\end{figure*}

\section{Related Works}
\label{sec:related}
Cross-modal keypoint matching can take various forms~\cite{jiang2021review}.
Some are designed for multi-spectral settings, such as matching near-infrared images with visible-light images~\cite{baruch2021joint}. 
Others focus on visible-to-infrared matching~\cite{tuzcuouglu2024xoftr} or RGB-to-depth maps or satellite imagery~\cite{jiang2021review}.
Arguably, temporal variations~\cite{verdie2015tilde}, where the same scene is observed at different times of the day or year, can also be considered a type of multimodality when dealing with extreme cases.  
A further distinction can be made with medical imaging modalities due to the inherently dynamic and heterogeneous nature of tissue appearance~\cite{heinrich2012mind,juvekar2023remind}. The visibility of anatomical structures such as parenchyma, tumors, bones, fluids, and vessels, as well as functionally relevant tissues such as gray matter, white matter, and fiber tracts, varies significantly across imaging modalities (e.g., MRI, CT, iUS, PET, fMRI, or SPECT), posing unique challenges for descriptor design used for matching that need to generalize beyond appearance and toward shared structural, functional or semantic information~\cite{dey2024learning,dorent2023unified}.

\vspace{5pt}
\noindent\textbf{Multimodal Matching of 2D Medical Images:}
Methods addressing multimodal matching of medical images were initially developed for retinal imaging for fundus-FA or fundus-OCT registration. Early handcrafted descriptors like PIIFD~\cite{chen2010partial} have since been outperformed by learning-based methods~\cite{lee2019deep,liu2022semi,santarossa2022medregnet}, offering enhanced robustness to intensity variations, rotation, and sparse annotations through end-to-end keypoint learning.
In~\cite{grewal2020end}, an end-to-end self-supervised Siamese CNN was proposed for detecting and matching anatomical landmarks in pairs of 2D lower abdominal CT slices. The network jointly learns keypoint locations and descriptors, achieving high-density correspondences under intensity, affine, and elastic transformations. 
Diffusion-guided image registration leveraging features from off-the-shelf diffusion models was proposed in~\cite{tursynbek2025guiding}. The model was pretrained on natural RGB images as a semantic similarity measure for deformable matching. Applied to both multimodal 2D Dual-energy X-ray to X-ray and monomodal MRI (2D slices) matching, their approach enables anatomically meaningful alignment even in cases of missing anatomy.
Recently, we proposed a Siamese architecture based on a contrastive learning strategy~\cite{rasheed2024learning} to learn to match 2D keypoints between preoperative MRI and iUS image.
This 2D method showed robustness to speckle appearance changes.
In parallel to keypoint descriptors, recent foundational models for multimodal matching have demonstrated impressive results in medical imaging. Approaches such as MINIMA~\cite{ren2025minima} and MatchAnything~\cite{he2025matchanything}, both based on Transformer architectures, leverage large-scale synthetic datasets to enable modality-invariant correspondence across a wide range of tasks, including intra-modality MRI (PD–T1, PD–T2, T1–T2), cross-modality structural and functional imaging (MRI–PET, CT–SPECT), and diverse modality pairs such as CT–MRI, PET–MRI, and fundus–OCT or fundus–FA, without requiring task-specific tuning. However, since most medical 2D images represent slices of 3D volumes rather than 2D projections of 3D scenes, these 2D keypoint methods are inherently limited as they assume that both images lie on the same anatomical plane in addition to not accounting for 3D deformations.

\vspace{5pt}
\noindent\textbf{Keypoints Matching for 3D Images and Volumes:}
Less attention has been given to three-dimensional data. Prior efforts to extend conventional 2D descriptors to 3D such as Harris~\cite{sipiran2011harris}, SIFT~\cite{rister2017volumetric}, SIFT-rank~\cite{chauvin2020neuroimage} and SURF~\cite{agier2016hubless} have highlighted unique challenges in 3D, where orientation, sampling, and viewpoint changes become more complex, increasing computational cost.
More advanced approaches were therefore proposed, combining, for example, the F\"orstner operator with Normalized Gradient Fields (NGF) to detect and describe 3D keypoints in CT images~\cite{ruhaak2017estimation}, with successful use in registration. 
For CT matching, recent deep learning methods have shown that self-supervised training on synthetic 3D patch deformations~\cite{grewal2023automatic} or affine transformations~\cite{loiseau2021learning} significantly improves matching and registration performance.
One of the first multimodal 3D descriptors, MIND~\cite{heinrich2012mind}, leverages patch-based self-similarity to enable deformable multimodal registration, originally for MRI–CT and more recently for MRI–iUS.
Recently, KeyMorph~\cite{evan2021keymorph}, and its foundational variant BrainMorph~\cite{wang2024brainmorph}, tackled varying MRI contrasts by learning anatomically semantic keypoints in an unsupervised manner and computing transformations in closed form, enabling robust and interpretable alignment. Evaluated on brain MRI with varying contrasts (T1, T2, and PD-weighted), the method improves performance under large misalignments and across various contrast pairs.
2D/3D keypoint-based methods have also been proposed for multimodal volume-to-image registration using learned, detector-free features. Notable examples include automatic registration of X-ray to CT images~\cite{esteban2019towards} and freehand iUS to preoperative MRI volumes~\cite{markova2022global}. These approaches eliminate the need for manual initialization typically required by optimization-based methods and are robust to limited or noisy training data, making them particularly suitable for intraoperative settings.

\added{Most existing work on keypoint matching is complementary to image registration, while addressing a distinct subproblem. Keypoint matching aims to establish sparse but reliable correspondences, whereas image registration integrates these correspondences globally through optimization and regularization. In our framework, matching precision and spatial coverage directly influence downstream registration robustness, particularly in the presence of large cross-modal appearance changes. This motivates our focus on improving cross-modal correspondences as a prerequisite for reliable registration.}\rev{R1}

\section{Methods}
\label{sec:methods}

\subsection{Problem Formulation, Challenges and Strategy}

\textbf{Problem formulation:} Our method detects, describes and predicts a partial assignment between two sets of keypoints extracted from 
a 3D pre-operative MR volume $I_{\text{MR}} \in \mathbb{R}^{\Omega}$ and a 3D iUS volume $I_{\text{US}} \in \mathbb{R}^{\Omega}$, where $\Omega$ denotes the spatial domain. Each keypoint $i$ is composed of a 3D point position $\mathbf{p}_{i} =(x,y,z) \in [0,1]^3$, normalized by the image size, and a descriptor $\mathbf{d}_{i} \in \mathbb{R}^d$ that characterizes the local 3D information. Images $I_{\text{MR}}$ and $I_{\text{US}}$ have $N$ and $M$ keypoints, independently detected.
Given these two sets of keypoints, we seek a partial assignment matrix $\mathbf{P}\in \{0,1\}^{N\times M}$ between keypoints in $I_{\text{MR}}$ and $I_{\text{US}}$. 
Each keypoint can be matched at most once, as it originates from a unique 3D position, and some keypoints cannot have valid correspondences, due to occlusion or non-repeatability. The assignment matrix $\mathbf{P}$ is thus sparse.

\textbf{Challenges:} Current methods typically require large amounts of paired training data with known correspondences to learn robust and generalizable keypoints. However, applying such general-purpose models out-of-the-box to new modalities or clinical scenarios, such as MR and iUS volumes, remains challenging, as their performance heavily depends on exposure to sufficient representative data during training. In particular, acquiring accurate 3D correspondences between MR and iUS volumes demands rare clinical expertise and is time-consuming, limiting the feasibility of building such large paired datasets. An alternative is to train deep keypoint detectors and descriptors on co-registered MR-iUS volume pairs. 
Yet, even when large-scale datasets are available, they often lack the precise MR-iUS co-registration necessary for accurate modeling, making this a persistent bottleneck.

\textbf{A patient-specific, matching-by-synthesis strategy:} 
Instead of designing a general framework to build $\mathbf{P}$, we propose to design a patient-specific approach for detecting, describing and matching keypoints between iUS and preoperative MR volumes. Patient-specific training has demonstrated superior performance over patient-agnostic approaches in cases where the preoperative image is informative and the intraoperative image can be reliably synthesized \cite{gopalakrishnan2025rapid,fehrentz2024intraoperative,dorent2024patient}.
In our context, we propose to leverage synthetic iUS volumes generated from preoperative MRI to 1) identify keypoints that are salient and common across modalities; 2) describe the local information in a cross-modal manner using contrastive learning; and 3) match the most discriminative correspondences. At inference, our trained approach is used to identify a set of correspondences between preoperative MRI and real iUS images.
Figure \ref{fig:overview} illustrates our method.

\rev{R1, R2}
\begin{figure}[h!]
     \begin{center}
        \includegraphics[width=1\linewidth]{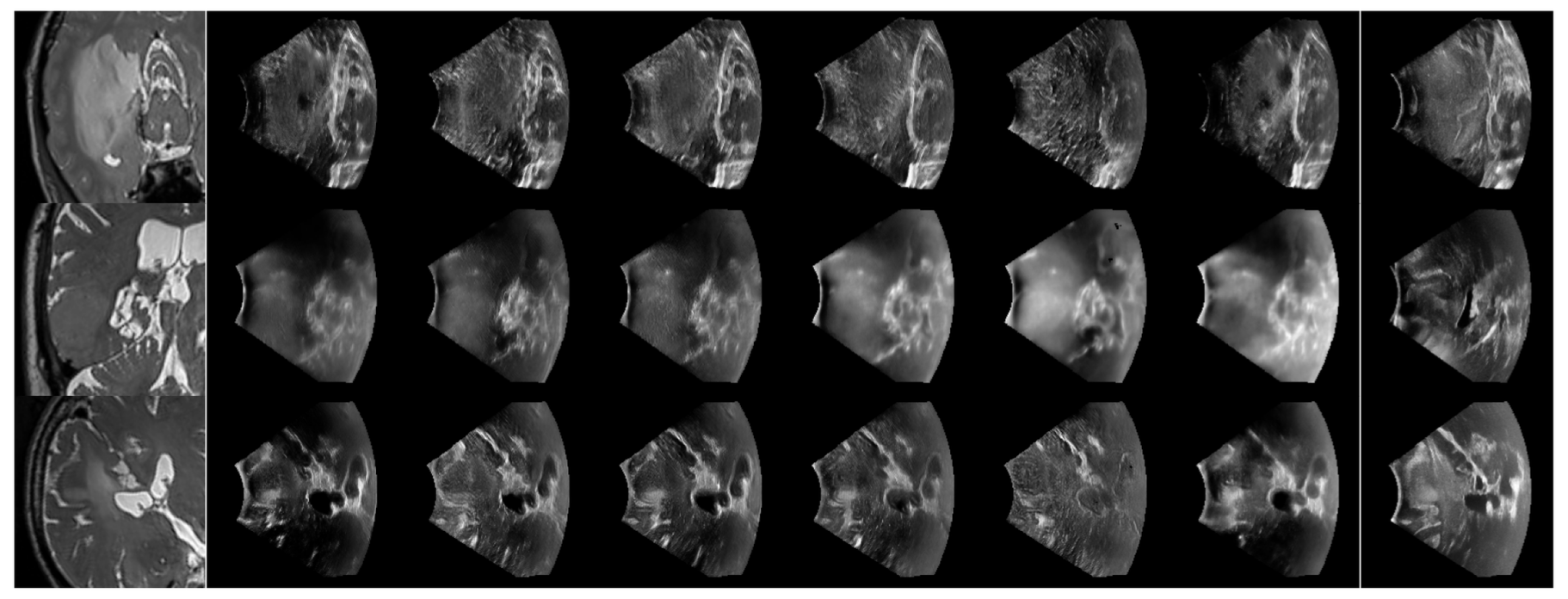}
    \end{center}
    \caption{\added{Synthetic US image generations for three different T2 MR images (One case per row) using MMHVAE~\cite{dorentUnifiedCrossModalImage2024}. The first column shows T2 MR; the middle columns show samples of synthetic US images generated using different combinations of T2, T1, and FLAIR with different speckles; the last column shows the ground truth US image.}}
    \label{fig:synth}
\end{figure}

\subsection{Creating the Patient-Specific Training Set}
\label{sec:image_synthesis}
Since iUS images cannot be acquired prior to brain surgery, we propose to construct a paired training set of preoperative MR and iUS volumes by synthesizing iUS from MR images. To this end, we leverage the recently proposed MMHVAE framework~\cite{dorentUnifiedCrossModalImage2024}, a hierarchical variational auto-encoder designed for incomplete multimodal data, to generate realistic iUS from preoperative MRI. This framework is particularly well-suited to our task because 1) it can handle incomplete multi-parametric MR inputs, a frequent scenario in clinical settings, and 2) it enables the generation of synthetic iUS images with diverse appearances (e.g., varying texture and speckle patterns) by adjusting the scale parameter $\gamma$ of the standard deviations within its hierarchical latent structure. 
\added{Figure~\ref{fig:synth} provides qualitative comparisons between real and synthetic intraoperative ultrasound volumes, illustrating structural realism across modalities.}\rev{R2}

To build a large and diverse dataset that encourages generalization to real US, we exploit both of these properties. Specifically, let us assume we have access to $K$ pre-operative MR sequences among T1, T2 and T2-FLAIR, we synthesize iUS for each of the $2^K-1$ combinations of MR sequences and for multiple values of the scale parameter $\gamma \in \{0.3, 0.5, 0.7, 1\}$. This leads to the creation of a dataset of synthetic iUS images with different appearances denoted as $\mathcal{D}_{\text{synUS}}:=\{ I^{(k)}_{\text{synUS}}\}_{k=1}^{4\times(2^K-1)}$.

\added{The MMHVAE synthesis module is trained once as a general model on patients distinct from those used for descriptor learning and registration, and is not retrained per patient. This separation enables cross-patient generalization at the synthesis level while preserving patient-specific optimization for matching.}\rev{R3}

\subsection{Keypoint Detection and Sampling Strategy}
\label{sec:keypoint_detection} We seek to identify keypoints that serve two complementary objectives: 1) sampling pairs of positive and negative correspondences between MR and iUS to train a deep cross-modal descriptor with effective patch-based augmentation, and 2) providing a spatial prior during inference to select informative keypoints in the pre-operative MR image. To fulfill these goals, keypoints must satisfy three main criteria: (i) be located in salient regions to ensure discriminativeness, (ii) be consistent across modalities for efficient cross-modal matching, and (iii) be spatially diverse to enhance training variability and avoid false negative pairs. To this end, we introduce a novel stochastic detection strategy based on a probabilistic cross-modal saliency map.

\subsubsection{Initial independent detection} We begin by detecting salient keypoints independently in the preoperative MR volume $I_{\text{MR}}$ and in the synthetic iUS dataset $\mathcal{D}_{\text{synUS}}$ using SIFT3D~\cite{risterVolumetricImageRegistration2017}, a volumetric extension of the classical SIFT algorithm. While SIFT3D is effective at detecting texture-rich regions, its sensitivity to modality-specific appearance, such as the differences between MR and iUS, result in poor consistency between cross-modal detections. This motivates the need for a joint keypoint detection strategy that accounts for saliency across modalities.

\subsubsection{Constructing cross-modal saliency heatmaps} We propose a cross-modal saliency heatmap by aggregating spatial statistics of SIFT3D keypoints across MR and synthetic iUS pairs. Specifically, for each synthetic iUS volume $I^{(k)}_{\text{synUS}} \in \mathcal{D}_{\text{synUS}}$, we compute a descriptor presence mask $M^{(k)}_{\text{synUS}} \in \{0,1\}^{\Omega}$, where $M^{(k)}_{v, \ \text{synUS}} = 1$ if a SIFT3D keypoint is detected at voxel $v$, and 0 otherwise. We then construct a heatmap $P_{\text{US}}$ by summing the presence masks over the full synthetic dataset, i.e. $P_{\text{US}} = \sum_{k=1}^{4 \times (2^K - 1)} M^{(k)}_{\text{synUS}}$; applying a Gaussian smoothing filter ($\sigma=2$) to $P_{\text{US}}$, then normalizing the resulting volume to the range $[0, 1]$. This heatmap highlights salient regions stable across synthetic iUS appearances.

To account for salient regions in pre-operative MRI, a heatmap $P_{\text{MR}}$ is similarly constructed by smoothing and normalizing the descriptor presence mask of $I_{\text{MR}}$. 
We then fuse the modality-specific saliency maps into a joint saliency map $P_\text{comb}$ using a probabilistic OR operation:
\mbox{$P_{\text{comb}}=1-(1-P_{\text{MR}})(1-P_{\text{US}})$}, increasing the likelihood of a voxel being considered discriminative if it is salient in at least one modality, and even more so when it is salient in both.

\subsubsection{Applying a spatial prior}
To constrain sampling to clinically relevant regions and avoid off-field points in iUS, we apply a spatially-weighted FoV mask $M_w$. This mask is constructed by computing the Euclidean distance from each voxel to the center of mass of the iUS FoV, followed by Gaussian smoothing to emphasize central regions. The resulting residual saliency map is $P_{\text{res}}=P_{\text{comb}}\cdot M_w$

\subsubsection{Sampling training keypoints}
During training, keypoints are randomly sampled from the residual saliency map $P_{\text{res}}$ via a sequential rejection sampling procedure. The sampling procedure is designed to enforce spatial and FoV coverage constraints. Specifically, the goal is to obtain a set of $N$ keypoints satisfying the following conditions: C1) For each keypoint $i$, at least $80\%$ of the corresponding patch centered at $\mathbf{p}_i$ lies within the iUS FoV;
C2) The Euclidean distance between any pair of selected keypoints $\mathbf{p}_i$ and $\mathbf{p}_j$ (with $i\neq j$) is at least $2\text{mm}$, ensuring spatial diversity and preventing false negative correspondences (see Figure~\ref{fig:overview}).

\begin{figure}[t!]
    \centering
    \includegraphics[trim=0 0 0 0, clip, width=\linewidth]{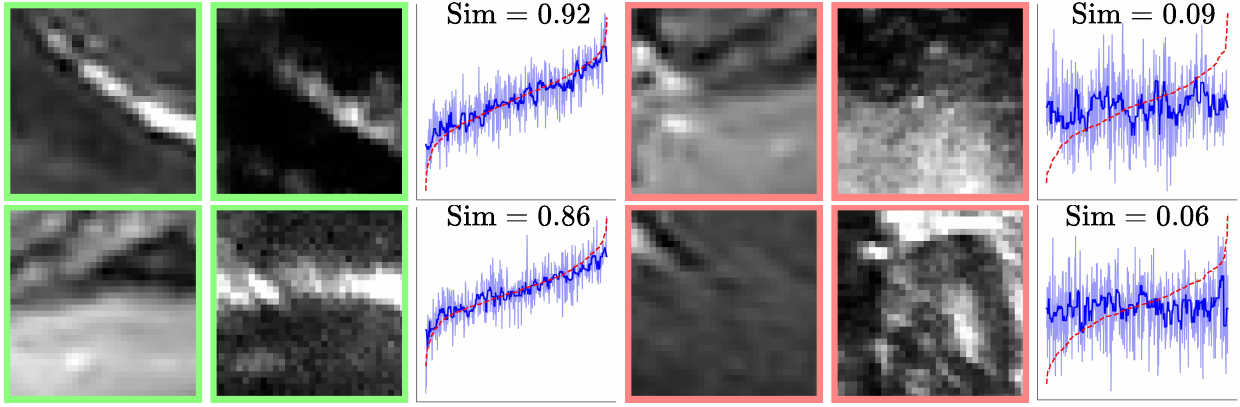}
    \caption{
Examples of MR-iUS patches showing high descriptor similarity for positive pairs (left) and low descriptor similarity (right) for negative pairs. The $d$-dimensional feature vectors were sorted according to the values of the MR descriptor $\mathbf{d}^{\text{MR}}$, highlighting the correlation between MR and iUS descriptors.}
    \label{fig:patches}
\end{figure}

\subsection{Cross-Modal 3D Feature Descriptor}
\label{sec:descriptor} 
To design the cross-modal 3D descriptor, we employ a contrastive learning approach based on a Siamese deep learning architecture that maps patches from the MRI or iUS domain, to a $d$-dimensional feature space. This shared feature space is designed to ensure that patches centered at corresponding anatomical locations in MRI and iUS produce similar descriptors and that feature descriptors are anatomically discriminative, to enable differentiation between distinct anatomical regions.

\subsubsection{Patch extraction and feature descriptors}
Given a set of $N$ keypoints, we extract cubic patches of size $s^3$ centered at each location $\mathbf{p}_i$. Since the pre-operative MR and synthetic iUS volumes are spatially co-registered, positive training pairs $(\mathbf{v}^{\text{MR}}_i,\mathbf{v}^{\text{US}}_i)$ can be extracted, consisting of an anchor MR patch $\mathbf{v}^{\text{MR}}_i$ and a corresponding iUS patch $\mathbf{v}^{\text{US}}_i$ from a randomly selected synthetic iUS volume. These patches are fed to a shared 3D ResNet-18 encoder $\Phi: \mathbb{R}^{s^3} \rightarrow \mathbb{R}^d$, which produces L2-normalized feature descriptors $\mathbf{d}^{\text{MR}}_i = \Phi(\mathbf{v}^{\text{MR}}_i)$ and $\mathbf{d}^{\text{US}}_i = \Phi(\mathbf{v}^{\text{US}}_i)$ for the MR and iUS modalities, respectively.

To construct negative pairs $(\mathbf{v}^{\text{MR}}_i, \mathbf{v}^{\text{US}}_n)$, we leverage our keypoint sampling strategy, which promotes spatial diversity and limits spatial redundancy. Specifically, negative iUS patches are sampled at locations $\mathbf{p}_i$ and $\mathbf{p}_j$, such that $j \ne i$.

\subsubsection{Hard negative mining strategy} Different training losses, such as Binary Cross-Entropy (BCE), Noise-Contrastive Estimation (infoNCE), and triplet loss, can be used to learn cross-modal feature descriptors. We empirically chose a triplet loss
that encourages negative pairs of descriptors $(\mathbf{d}^{\text{MR}}_i, \mathbf{d}^{\text{US}}_n)$ to be distant from any positive pairs $(\mathbf{d}^{\text{MR}}_i, \mathbf{d}^{\text{US}}_i)$ by at least a certain margin value $m$ and is defined as:
\begin{equation}
    \mathcal{L}_{\text{triplet}} = \max\left(0, \|\mathbf{d}^{\text{MR}}_i - \mathbf{d}^{\text{US}}_i\|_2^2 - \|\mathbf{d}^{\text{MR}}_i - \mathbf{d}^{\text{US}}_n\|_2^2 + m\right)
\end{equation}

A critical component of effective triplet loss training is the selection of informative negative examples. Without a hard negative mining strategy, training rapidly saturates, as many triplets contribute negligible gradient~\cite{hermans2017defense}. To address this, we propose a progressive hard negative mining scheme grounded in curriculum learning: at early stages of training, negatives are sampled from spatially distant regions, which are expected to be easier to be distinguished in the feature space. As training progresses and the model improves, the mining progressively shifts toward harder negatives, which corresponds to negative descriptors that are closer in feature space (See Figure \ref{fig:patches}).

At each training step, for a given anchor descriptor $\mathbf{d}^{\text{MR}}_i$, we compute a selection score $S_{i,j}$ for all $j\neq i$ keypoints in the batch, as:
\begin{equation}
S_{i,j} = 
(1 - \lambda_t) \underbrace{\min \left(\frac{\| \mathbf{p}_i - \mathbf{p}_j \|_2}{D_{\max}}, 1 \right)}_{\text{spatial distance}}
- \lambda_t \underbrace{\| \mathbf{d}^{\text{MR}}_i - \mathbf{d}^{\text{US}}_i \|_2 \vphantom{\min \left(\frac{\| \mathbf{p}_i - \mathbf{p}_j \|_2}{D_{\max}}, 1 \right)}}_{\text{feature similarity}} \ ,
\end{equation}
where $\lambda_t \in [0,1]$ weights the normalized Euclidean distance bounded by $D_{\max}=24\text{ mm}$ and the feature similarity. The weight 
$\lambda_t$ acts as a curriculum difficulty scheduler that progressively shifts the emphasis of negative mining from anatomically distant pairs (high spatial weight) to semantically similar pairs (high feature similarity). In practice,  $\lambda_t$ is defined at epoch $t$ as: $\lambda_t = \min\left(t/T, 1.0\right)$,
where $T$ is the number of warm-up epochs, which was set to $200$ in our experiments.
For each anchor $i$, the candidate negative $j$ corresponds to the keypoint with the lowest score $S_{i,j}$.

\subsubsection{Encouraging rotation-invariance descriptors} To encourage rotation invariance in the learned descriptors, we apply random 3D rotations to the MR anchor patches during training. Following a curriculum learning strategy similar to our hard negative mining scheme, the rotation angle is uniformly sampled from the range $[0^\circ, \theta_{\max}]$, where $\theta_{\max}$ increases linearly from $0^\circ$ to $30^\circ$ over the first 1000 training epochs. To avoid information loss during rotation, we initially extract patches at 1.5 times the target size $s$ along each spatial dimension, and then center-crop them to obtain the final $s^3$ voxel input.

\subsubsection{Training details} At each epoch, one synthetic iUS $I^{(k)}_{\text{synUS}}$ and $1024$ keypoints are sampled using the strategy defined in Section~\ref{sec:keypoint_detection}. 
\added{We fix the number of sampled candidate keypoints to $N=1024$, which we found to be a practical trade-off between spatial coverage and computational cost. Performance was stable around this value, with larger $N$ increasing runtime without measurable accuracy gains.}\rev{R2}
We extract patches of size $s=32$ around each keypoint in both preoperative $I_{\text{MR}}$ and synthetic iUS $I^{(k)}_{\text{synUS}}$. We use a batch size of $N=256$, leading to $4$ iterations per epoch. Online negative mining is performed within the batch to construct hard triplets dynamically. The network is trained for $2000$ epochs using AdamW with an initial learning rate of $ 10^{-3}$, weight decay of $2 \times 10^{-3}$, and a cosine annealing schedule with a minimum learning rate of $10^{-6}$. 

\subsection{Keypoint Matching at Testing Time} At test time on a real iUS volume, we construct a sparse assignment matrix $\mathbf{P} \in \{0,1\}^{N \times M}$ using a nearest neighbor matching strategy followed by an ambiguity filtering step. Specifically, we sample $N = 1024$ keypoints in the MR volume using the procedure described in Section~\ref{sec:keypoint_detection}. Since no saliency map is available for the real iUS image, a uniform 3D grid of step size $4$ mm within the FoV of the real iUS volume is used to define $M$ keypoints.
This allows for unbiased coverage of the visible anatomy while avoiding detecting modality-specific salient structures.
For each MR keypoint $i$, we identify its best matching iUS keypoint $j$ by comparing their descriptors $\mathbf{d}^{\text{MR}}_i$ and $\mathbf{d}^{\text{US}}_j$ using the $L_2$ norm. Lowe’s ratio test is used to filter out unreliable correspondences.

\section{Results}
\label{sec:results}
To assess our approach, we conducted comprehensive experiments on matching and registration between MRI and iUS in patients with brain tumors. We evaluate performance through: (i) paired image matching using standard paired evaluation metrics, (ii) an ablation study analyzing each novel component and a robustness analysis to rotation, and  (iii) image registration aligning pre-operative MRI with post-resection iUS.

\subsection{Dataset}
We conducted our experiments using the Brain Resection Multimodal Imaging Database (ReMIND)~\cite{remind}, which contains pre-operative multi-parametric MRI and iUS images from 114 consecutive patients. For this study, we focused on the subset of 13 patients for whom complete pre-operative 3D MRI (T1, T2, and FLAIR) and pre-dural iUS were available. The iUS volumes were reconstructed from tracked 2D handheld probe acquisitions. All images were resampled to an isotropic resolution of $0.5$ mm, zero-padded to an in-plane size of $192 \times 192$, and intensity-normalized to the $[0, 1]$ range.

To evaluate cross-modal image matching, we constructed a paired dataset by co-registering pre-dural iUS volumes with pre-operative MRI following the protocol described in~\cite{dorentUnifiedCrossModalImage2024}. \added{During method development, 2 cases were randomly selected for parameter tuning and prototyping. The finalized approach was then evaluated on the remaining 11 test cases. We recall that a separate model is trained for each case (patient-specific) and models are not trained for inter-patient generalization.}\rev{R1}

For the downstream registration task, we used the 4 validation cases from the ReMIND2Reg challenge 2024~\cite{dorent_2024_10991880, dorent2025brain}, which provide manually annotated paired landmarks as ground truth.

\begin{table*}[ht]
\centering
\caption{Per-case performance across state-of-the-art methods. Metrics reported are Precision (P. \%), Matching Score (MS \%), and total matched points (MP). Matching Scores for 2D methods are not reported since they are detector-free.}
\resizebox{1\textwidth}{!}{
\begin{tabular}{l*{9}{ccc}}
\toprule
\textbf{Case} & \multicolumn{3}{c}{\textbf{SIFT3D}} & \multicolumn{3}{c}{\textbf{Förstner+NGF}} & \multicolumn{3}{c}{\textbf{MIND}} & \multicolumn{3}{c}{\textbf{MedicalNet}} & \multicolumn{2}{c}{\textbf{ALIKED}} & \multicolumn{2}{c}{\textbf{SuperPoint}} & \multicolumn{2}{c}{\textbf{LoFTR}} & \multicolumn{2}{c}{\textbf{LM2KD}} & \multicolumn{3}{c}{\textbf{Ours}} \\
\cmidrule(lr){2-4} \cmidrule(lr){5-7} \cmidrule(lr){8-10} \cmidrule(lr){11-13} \cmidrule(lr){14-15} \cmidrule(lr){16-17} \cmidrule(lr){18-19} \cmidrule(lr){20-21} \cmidrule(lr){22-24}
& P. & MS & MP & P. & MS & MP & P. & MS & MP & P. & MS & MP & P. & MP & P. & MP & P. & MP & P. & MP & P. & MS & MP \\
\midrule
\rowcolor{LightGray}
C1  & 48.0 & 3.19 & 68  & 1.3 & 0.04 & 152 & 0.8 & 0.08 & 99   & 0.7 & 0.05 & 73   & 10.5 & 2.5\,\text{k}    & 77.5 & 271  & 33.0 & 4.2\,\text{k} & 46.5 & 172  & 64.5 & 5.78 & 92 \\
C2  & 26.5 & 0.88 & 34  & 0.8 & 0.04 & 133 & 2.7 & 0.56 & 209  & 4.5 & 0.46 & 104  & 4.3 & 1.7\,\text{k}    & 57.9 & 394  & 24.1 & 2.6\,\text{k} & 35.8 & 81   & 64.2 & 5.00 & 80 \\
\rowcolor{LightGray}
C3  & 47.4 & 0.98 & 21  & 0.5 & 0.02 & 190 & 2.3 & 0.54 & 238  & 1.1 & 0.06 & 60   & 6.7 & 2.4\,\text{k}   & 93.3 & 2.7\,\text{k} & 54.4 & 5.1\,\text{k} & 62.9 & 35   & 66.3 & 3.60 & 55 \\
C4  & 43.2 & 1.69 & 41  & 1.3 & 0.06 & 309 & 0.6 & 0.11 & 188  & 0.5 & 0.03 & 59   & 13.0 & 3.1\,\text{k}   & 64.8 & 364  & 41.9 & 6.9\,\text{k} & 79.8 & 114  & 73.9 & 5.30 & 74 \\
\rowcolor{LightGray}
C5  & 48.3 & 1.42 & 30  & 0.4 & 0.02 & 241 & 1.2 & 0.32 & 279  & 0.8 & 0.04 & 55   & 4.7 & 2.8\,\text{k}    & 79.5 & 809  & 53.7 & 5.5\,\text{k} & 63.4 & 112  & 66.7 & 6.50 & 100 \\
C6  & 73.0 & 4.67 & 66  & 0.0 & 0.00 & 148 & 0.3 & 0.07 & 207  & 1.4 & 0.07 & 48   & 7.7 & 1.8\,\text{k}    & 3.3  & 30   & 33.3 & 1.9\,\text{k} & 90.8 & 76   & 80.2 & 3.90 & 50 \\
\rowcolor{LightGray}
C7  & 46.4 & 2.35 & 52  & 0.0 & 0.00 & 145 & 0.1 & 0.03 & 215  & 0.3 & 0.02 & 62   & 4.1 & 1.5\,\text{k}    & 29.0 & 62   & 46.0 & 4.9\,\text{k} & 66.7 & 72   & 58.4 & 2.30 & 40 \\
C8  & 57.3 & 1.90 & 34  & 0.0 & 0.00 & 195 & 2.1 & 0.39 & 188  & 1.0 & 0.06 & 59   & 7.5 & 3.0\,\text{k}    & 72.3 & 376  & 57.6 & 3.5\,\text{k} & 58.8 & 160  & 73.7 & 5.90 & 81 \\
\rowcolor{LightGray}
C9  & 68.2 & 5.18 & 78  & 2.0 & 0.12 & 150 & 0.5 & 0.11 & 226  & 2.6 & 0.17 & 66   & 5.0 & 1.2\,\text{k}    & 45.6 & 228  & 35.4 & 3.1\,\text{k} & 56.0 & 25   & 72.1 & 6.40 & 90 \\
C10 & 55.2 & 2.40 & 45  & 0.0 & 0.00 & 184 & 0.5 & 0.10 & 216  & 1.0 & 0.08 & 76   & 9.3 & 1.8\,\text{k}    & 55.9 & 152  & 34.3 & 4.1\,\text{k} & 15.9 & 63   & 74.9 & 3.50 & 47 \\
\rowcolor{LightGray}
C11 & 27.5 & 0.44 & 17  & 0.0 & 0.00 & 198 & 0.9 & 0.31 & 356  & 0.9 & 0.06 & 70   & 5.4 & 2.0\,\text{k}   & 2.4  & 125  & 39.9 & 3.2\,\text{k} & 46.2 & 106  & 73.0 & 6.20 & 87 \\
\midrule
\textbf{Mean} & 49.2 & 2.28 & 44  & 0.6 & 0.03 & 186 & 1.1 & 0.24 & 220  & 1.3 & 0.10 & 66   & 7.1 & 2.2\,\text{k}    & 52.9 & 502  & 41.2 & 4.1\,\text{k} & 56.6 & 92   & 69.8 & 4.90 & 72 \\
\rowcolor{LightGray}
\textbf{SD}  & 14.4 & 1.52 & 20  & 0.7 & 0.04 & 52  & 0.9 & 0.19 & 63   & 1.2 & 0.13 & 15   & 2.9  & 643    & 30.2 & 764  & 10.6 & 1.4\,\text{k} & 20.5 & 46   & 6.3  & 1.40 & 21 \\
\bottomrule
\end{tabular}
}
\label{tab:casewise_metrics_9methods}
\end{table*}

\begin{figure*}[h!]
\centering
\includegraphics[width=1\linewidth]{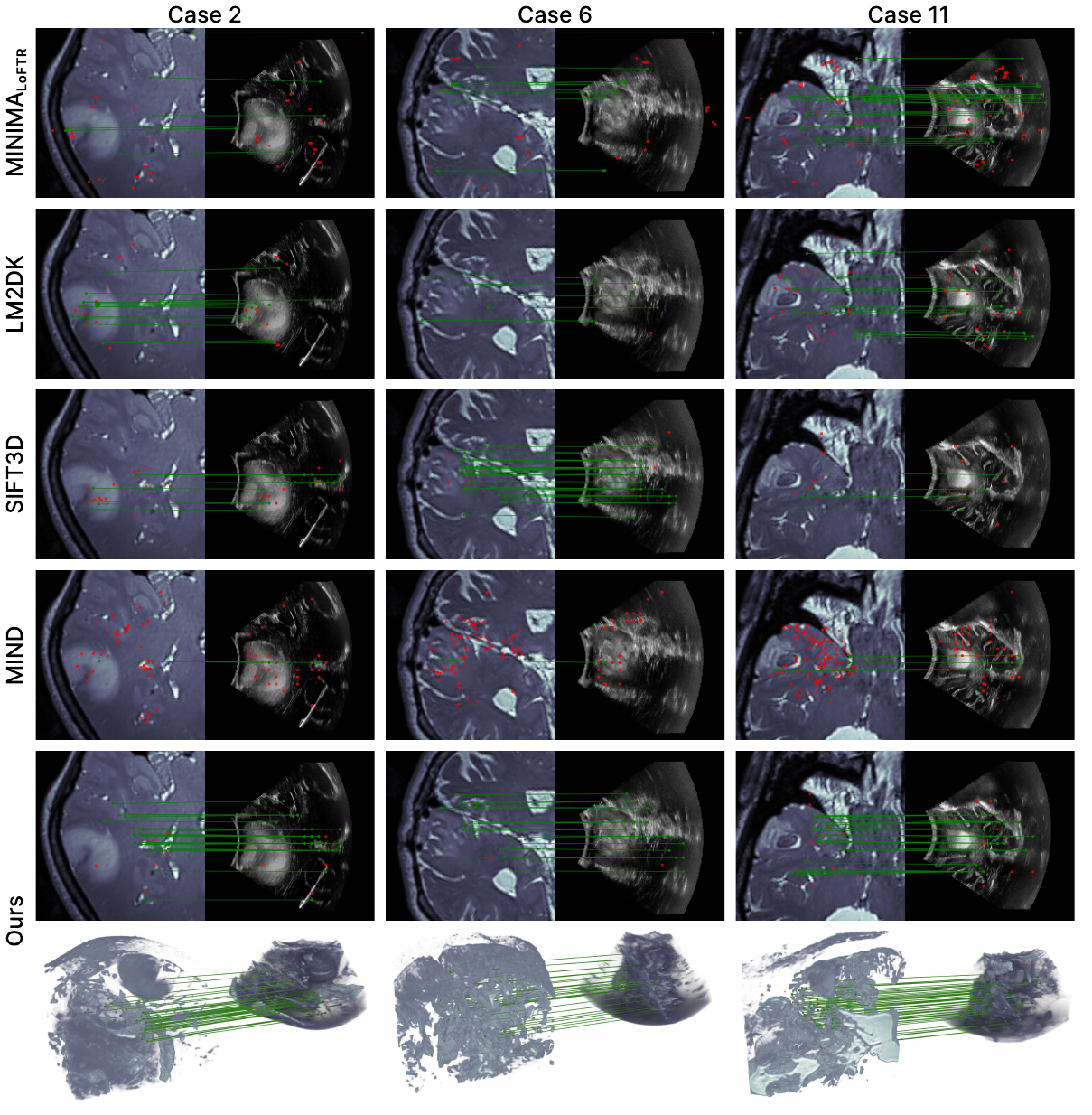}
\caption{Qualitative matching results across three cases (columns).  Rows 1–5 show results on slices from the 5 best-performing methods. Green lines indicate correct matches; red dots denote mismatches. Last row shows volume rendering with matching using our descriptor.}
\label{fig:matching_results}
\end{figure*}

\subsection{Image Matching}
We evaluate the ability of our cross-modal matching approach to identify anatomically corresponding keypoints across modalities. Experiments are performed on the $11$ test cases using their pre-operative T2-weighted MRI and real pre-dural iUS images, with ground-truth correspondences obtained using 3D co-registration~\cite {dorentUnifiedCrossModalImage2024}. 

\subsubsection{Metrics} A match is considered correct if it lies within a predefined spatial tolerance ($2.5$ mm) of the transformed ground-truth location. 
Keypoint matching performance is evaluated using 1) Precision, as the proportion of correctly matched pairs, 2) Matching Score, as the proportion of correct matches relative to the number of detected keypoints in the MR volume, and 3) the number of established Matched Points.

\subsubsection{Comparison with Related Work}
We benchmarked our proposed feature descriptor (\textit{Ours}) against eight competing approaches. Four 2D matching approaches: \textit{SuperPoint} \cite{detone2018superpoint}, \textit{ALIKED} \cite{zhao2023aliked}, and \textit{LoFTR}\cite{tuzcuouglu2024xoftr} \added{(as a proxy to the 2D/3D slice-to-volume search approach \cite{markova2022global})}\rev{R1} that we paired with the multimodal matcher \textit{MINIMA}~\cite{ren2025minima}, and the patient-specific method by Rasheed et al. \cite{rasheed2024learning}, that we will refer to as \textit{LM2DK}; and four 3D descriptor approaches: \textit{SIFT3D}~\cite{rister2017volumetric},  \textit{MIND}~\cite{heinrich2012mind}, F\"orstner+NGF~\cite{ruhaak2017estimation} and MedicalNet~\cite{alzubaidi2021mednet}, a pre-trained CNN for medical imaging tasks (equivalent to ImageNet).
\added{All classical baselines (MIND, SIFT3D, F\"orstner+NGF) were evaluated using their standard formulations without retraining and paired with the same candidate sampling strategy to ensure a fair comparison.}\rev{R1}

\textbf{Implementation details.} 2D methods required adaptation for volumetric data. \textit{SuperPoint}, \textit{ALIKED}, and \textit{LoFTR}, paired with MINIMA matcher, were applied slice-wise across the 3D volumes using their native detection mechanisms with matching performed on corresponding slices and aggregated metrics computed. \added{Each competing method is therefore evaluated in its most favorable and commonly used setting}\rev{R3}. \textit{LM2DK} was evaluated slice-wise (all vs all) using its original \textit{SuperPoint}-based detector and KNN matcher.

For baseline 3D methods, we paired \textit{SIFT3D} with our sampling-based detector, as its native keypoint detector failed to produce correspondences. Since \textit{MIND} provides only a descriptor, we similarly used our sampling detector. For \textit{MedicalNet}, we applied global average pool on the last layer of its pre-trained ResNet-18 architecture to obtain descriptors, and also used our sampling detector for keypoint detection.
\added{It is important to note that we evaluated SIFT3D as a keypoint detector for the 3D methods; however, it failed to produce any valid matches, motivating the use of dense sampling. We thus exclude it from Table \ref{tab:casewise_metrics_9methods}.}\rev{R2}

Due to the stochastic nature of our detection strategy, we repeated this evaluation protocol 10 times per subject and averaged the results across all subjects. For detector-based methods, the Lowe's ratio threshold parameter $l$ was optimized via grid search to maximize precision while ensuring a minimum of 40 matched points. This led to the selection of $l = 0.75$ for \textit{Ours}, $l = 0.8$ for \textit{LM2DK} and \textit{MedicalNet}, and $l = 0.9$ for  \textit{MIND}, \textit{SIFT3D} and F\"orstner+NGF.

\textbf{Results.}
As shown in Table~\ref{tab:casewise_metrics_9methods} and Figure~\ref{fig:matching_results}, our method consistently outperforms all baselines across precision and matching score. It achieves a matching precision of $69.8\%$, a matching score of $4.90\%$, and an average of $72$ correct correspondences, demonstrating both high discriminative ability and robustness to cross-modal variations. In contrast, traditional 3D methods like SIFT3D, MIND, and Förstner+NGF offer limited cross-modal utility, even when paired with our detector. MedicalNet shows particularly poor performance with only $1.3\%$ precision. Among 2D approaches, LM2DK performs best with relatively high precision ($56.6\%$) but much lower matching coverage, while ALIKED, SuperPoint, and LoFTR paired with MINIMA show moderate performance but cannot match the robustness of our specialized 3D approach. \added{Run times are reported in Table \ref{tab:runtime}.}\rev{R2}

\added{We qualitatively assessed sensitivity to key hyperparameters, including the triplet margin and curriculum schedule, and observed stable convergence across a reasonable range of values, indicating that the reported matching performance is not driven by sensitive parameter tuning.}\rev{R1}

\subsubsection{Ablation Study}

\begin{figure}[h!]
\centering
 \subfloat{\includegraphics[width=0.49\linewidth]{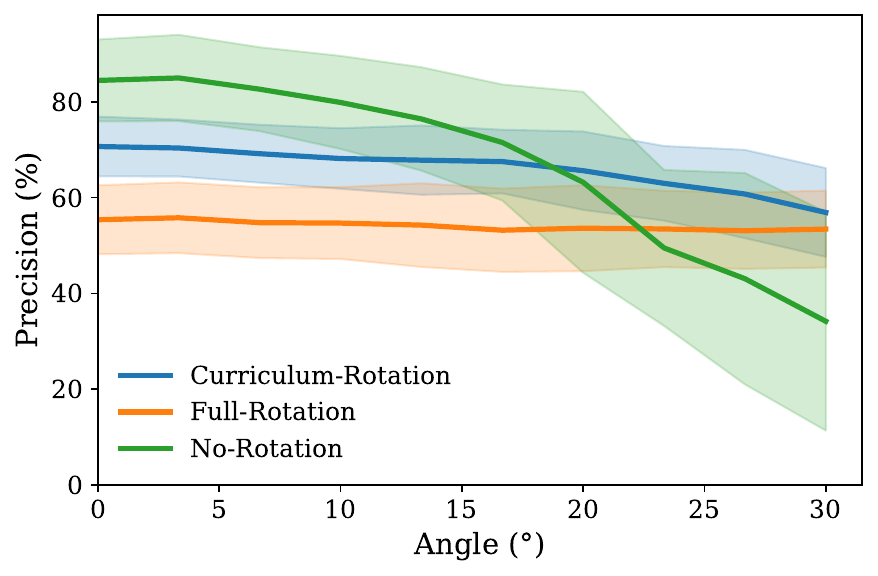}}
 \hfill
 \subfloat{\includegraphics[width=0.49\linewidth]{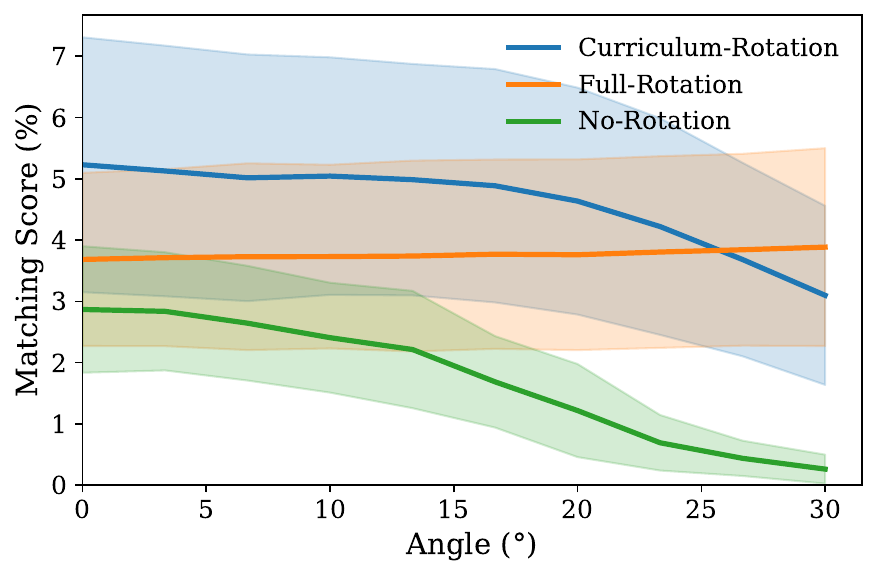}} \\
 \subfloat{\includegraphics[width=0.49\linewidth]{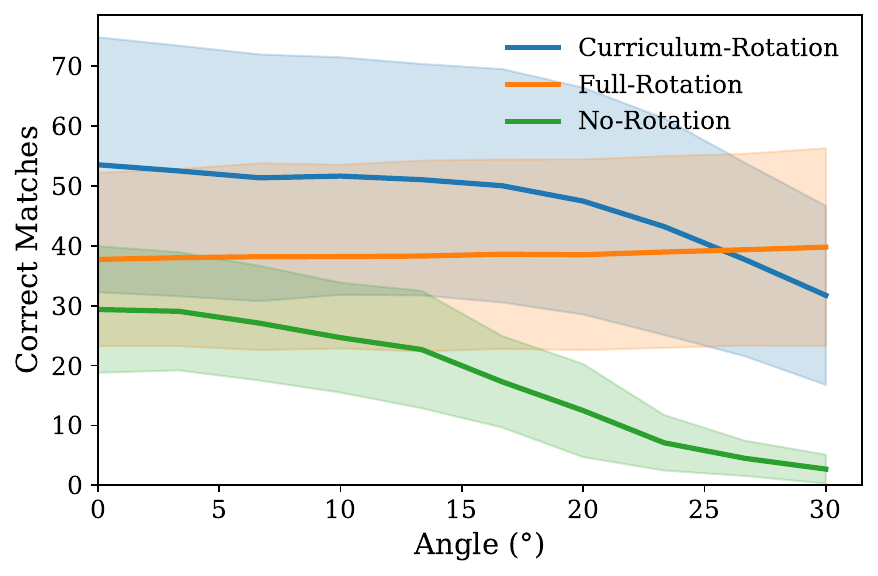}}
 \hfill
 \subfloat{\includegraphics[width=0.49\linewidth]{figs/rotation/rot_inv-correct_matches.pdf}}
\caption{Rotation invariance analysis across increasing rotation angles. Our curriculum-based model maintains higher matching quality under increasing rotational misalignment.}
\label{fig:rotation_invariance}
\end{figure}

We conduct a comprehensive ablation study to systematically evaluate four key design choices in our method: 1) synthetic training MR sequences, 2) keypoint detection and sampling strategy; 3) contrastive optimization objective; and  4) rotation invariance strategy. All components are evaluated on the keypoint matching task, with results summarized in Table~\ref{tab:ablation_study} and Figure~\ref{fig:rotation_invariance}. Our baseline configuration employs all synthetic modalities (T2, T1, and FLAIR), curriculum triplet loss, curriculum rotational augmentation, and stochastic keypoint detection.

\textbf{Impact of variation in iUS synthesis.} We investigate the contribution of exploiting different MR-derived synthetic US image by training from various combinations of T2, T1, and FLAIR sequences. T2 serves as the base modality in all configurations since it corresponds to our target MR sequence. Results in Table~\ref{tab:ablation_study} show the benefits of incorporating additional MR sequences during the synthesis of synthetic US images to increase the diversity of the iUS data. While individual addition of T1 or FLAIR improves performance over T2 only training, combining all three modalities yields optimal results, with the average number of matches increasing substantially from $18.9$ to $72.3$. This suggests that each modality contributes complementary anatomical information that enhances descriptor discriminability.

\rev{R2}
\begin{table}[h!]
\centering
\caption{\added{Training and inference runtime characteristics of evaluated descriptors.
For methods without learning, training time is not applicable (N/A).
Inference times are reported per image (2D) or per volume (3D).}}
\label{tab:runtime}
\begin{tabular}{lcc}
\toprule
\textbf{Method} & \textbf{Training Time} & \textbf{Inference Time} \\
\midrule
SIFT3D & N/A & $\sim$5000 ms / volume \\
F\"orstner+NGF & N/A & $\sim$300 ms / volume \\
MIND & N/A & $\sim$500 ms / volume \\
MedicalNet & Pretrained & $\sim$190 ms / volume \\
ALIKED & Pretrained & $\sim$10 ms / image \\
SuperPoint & Pretrained & $\sim$15 ms / image \\
LoFTR & Pretrained & $\sim$50 ms / image \\
LM2KD & $\sim$2 h / patient &  $\sim$10 ms / image \\ 
Ours & $\sim$5 h / patient & $<1000$ ms / volume \\
\bottomrule
\end{tabular}
\end{table}

\textbf{Stochastic vs deterministic detection.} We compare deterministic and stochastic keypoint sampling strategies. In the deterministic setting, a fixed set of SIFT3D keypoints is pre-selected from the pre-operative MRI ($I_{\text{MR}}$) and the synthetic dataset $\mathcal{D}_{\text{synUS}}$. In contrast, the stochastic approach dynamically samples keypoints during training. As reported in Table~\ref{tab:ablation_study}, stochastic sampling significantly increases model performance, particularly in precision ($69.79\%$ vs. $30.40\%$). This result suggests the benefit of exposing the model to a broader range of spatial contexts and more varied negative examples during training.

\textbf{Hard negative mining strategy.} To evaluate the effectiveness of our hard negative mining strategy combined with the triplet loss, we compared it against BCE and InfoNCE. As shown in Table~\ref{tab:ablation_study}, BCE yields the poorest performance, with a low precision of $8.6\%$, indicating limited discriminative capability. InfoNCE achieves a higher matching score ($6.2\%$) but at the cost of reduced precision ($60.4\%$), suggesting it is overly permissive in accepting matches. In contrast, our curriculum-based triplet loss offers the best trade-off across all metrics, attaining the highest precision ($60.4\%$), a correct matching score ($4.9\%$), and a sufficient number of correct matches ($72.3$), demonstrating its robustness in learning both discriminative and spatially meaningful representations.

\textbf{Curriculum rotation augmentation.} Finally, we evaluate the model’s robustness to rotations by comparing three training strategies: no rotational augmentation, full rotational augmentation (i.e., random rotations applied from the beginning) and our curriculum rotational augmentation. To assess rotation invariance, we test performance under 10 systematically increasing orientation discrepancies, ranging from $0^\circ$ to $30^\circ$ in $3^\circ$ increments, applied around $5$ randomly sampled 3D axes. As shown in Figure~\ref{fig:rotation_invariance}, the no-rotation baseline drops in performance as rotation increases, highlighting its inability to generalize to unseen orientations. The full-rotation model demonstrates more consistent performance across rotation angles but underperforms overall, suggesting that early exposure to high rotational variability may limit the capacity of the model to extract semantically meaningful features. In contrast, our curriculum-based approach obtains high performance up to moderate rotation angles ($<20^\circ$), with a gradual decline thereafter, demonstrating a more effective trade-off between robustness and discriminative capacity.

\def\arraystretch{1}
\begin{table}[t]
\centering
\caption{Ablation Study}
\begin{tabular}{p{2.5em} p{0.5em} p{2em} | c | c | c }
\toprule
 \multicolumn{3}{c|}{\textbf{Configurations}} & \textbf{Prec. (\%)} \hfill & \textbf{MSc. (\%)} \hfill &  \textbf{MP} \\ 
 
 \midrule
 
 \rowcolor{LightGray}
 \multicolumn{3}{l|}{Synth. Modalities} & & & \\ 
 \multicolumn{3}{l|}{\hspace{10pt}T2 \hspace{5pt}T1 FLAIR} &  &  &  \\
 
 \hspace{10pt}\, $\bullet$ & $\circ$ & $\circ$ & 68.47 $\pm$ 10.42 & 1.26 $\pm$ 0.26 & 18.94 $\pm$ 3.30 \\  
 \hspace{10pt}\, $\bullet$ & $\circ$ & $\bullet$ & 68.86 $\pm$ 5.94 & 3.68 $\pm$ 0.57 & 54.67 $\pm$ 6.20 \\  
 \hspace{10pt}\, $\bullet$ & $\bullet$ & $\circ$ & 69.86 $\pm$ 6.49 & 3.59 $\pm$ 0.57 & 51.94 $\pm$ 6.62 \\  
 \hspace{10pt}\, $\bullet$ & $\bullet$ & $\bullet$ &  \textbf{69.79} $\pm$ \textbf{4.94} & \textbf{4.92} $\pm$ \textbf{0.62} & \textbf{72.32} $\pm$ \textbf{7.92}   \\ 

 \midrule

 \rowcolor{LightGray}
 \multicolumn{3}{l|}{Optimization Loss} & & & \\ 
 \multicolumn{3}{l|}{\hspace{10pt}BCE} & 8.64 $\pm$ 1.00 & 4.20 $\pm$ 0.53 & \textbf{503.07} $\pm$ \textbf{18.31} \\
 \multicolumn{3}{l|}{\hspace{10pt}InfoNCE} & 60.39 $\pm$ 4.67 & \textbf{6.18} $\pm$ \textbf{0.70} & 104.13 $\pm$ 8.78\\
  \multicolumn{3}{l|}{\hspace{10pt}Triplet} & \textbf{69.79} $\pm$ \textbf{4.94} & 4.92 $\pm$ 0.62 & 72.32 $\pm$ 7.92 \\

 \midrule

 \rowcolor{LightGray}
 \multicolumn{3}{l|}{Point Detection} & & & \\ 
 \multicolumn{3}{l|}{\hspace{10pt}Deterministic} & 30.40 & 3.11 & \textbf{105.55} \\
 \multicolumn{3}{l|}{\hspace{10pt}Stochastic} & \textbf{69.79} $\pm$ \textbf{4.94} & \textbf{4.92} $\pm$ \textbf{0.62} & 72.32 $\pm$ 7.92 \\

 \bottomrule

\end{tabular}
\label{tab:ablation_study}
\end{table}

\subsection{Image Registration}
\label{subsec:registration_results}
We tested our method on the task of registering preoperative MR and post-resection iUS volumes using the publicly available ReMIND2Reg dataset, part of the Learn2Reg 2024 challenge. This dataset contains cases with large tissue deformation and topological changes due to tumor resection, and includes manually annotated ground-truth landmarks in both modalities, enabling quantitative assessment via Target Registration Error (TRE). 

\subsubsection{Keypoint-based iterative registration}
Our pipeline performs registration over three rigid alignment iterations. In each iteration, keypoint correspondences between the fixed MR volume and the current state of the moving iUS volume are computed using our descriptor.
These sparse matches are then used to estimate a rigid transformation via RANSAC, configured with a maximum of 4000 iterations and a 5.0~mm inlier threshold. The resulting transform is composed with previous transformations, and the original moving iUS image is resampled accordingly for the next iteration. The final output is the cumulative rigid transformation across all iterations.

\rev{R2,R3}
\begin{figure}[h!]
     \begin{center}
        \includegraphics[width=1\linewidth]{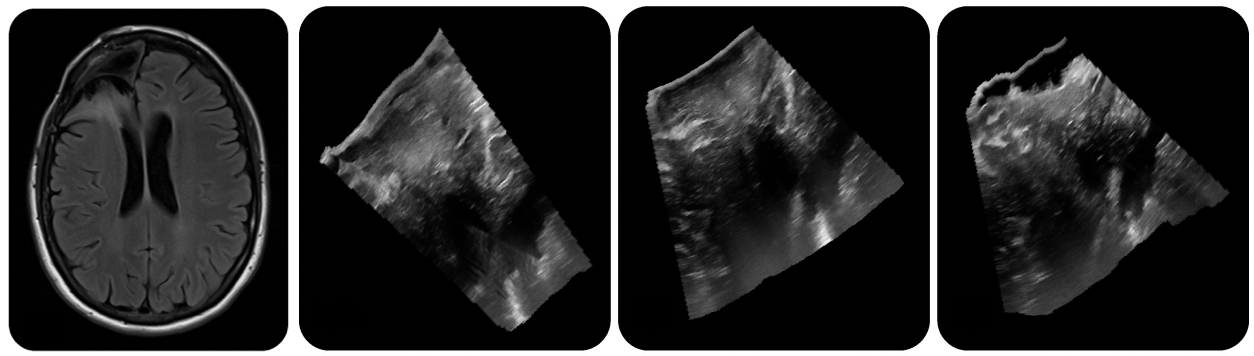}
    \end{center}
    \caption{\added{Illustrative example of one dataset from left to right: Preoperative T2-weighted MR; Intraoperative US prior to dural opening; Intraoperative US post dural opening; Intraoperative US prior to iMRI. Post-resection US images present important challenges for  registration algorithms due to large tissue deformation and topological changes (Courtesy of \cite{remind}).}}
    \label{fig:USs}
\end{figure}

\subsubsection{Competing registration methods}
We compare our method against top submissions from the ReMIND2Reg 2024  challenge's leaderboard: 1) the \textit{VROC (Variational Registration on Crack)} approach by Madesta \textit{et al.}, employs a two-stage rigid registration pipeline with conventional iterative optimization, using Gaussian-smoothed inputs, masking, and intensity thresholding, and sequentially optimizes NCC and NGF metrics; 2) the \textit{next-gen-nn} by Wang \textit{et al.}, an unsupervised method using a multilevel correlation balanced optimization strategy based on a MIND-SSC based feature extractor; 3) \textit{Topological Higher-Order MRF} by Li \textit{et al.}, a deformable registration framework based on a topological higher-order MRF with a multiscale optimization performed using a multi-resolution Quadratic Pseudo-Boolean Optimization strategy; and; \textit{Coarse-to-Fine Registration with Style Transfer} by Wang \textit{et al.}, that employs a coarse-to-fine strategy using 3D CycleGAN for unpaired imaging style transfer to convert iUS images into synthetic T1-style MR images to create a more unified signal distribution across modalities before registration followed by a NiftyReg registration.

\begin{figure*}[h!]
\centering
    \includegraphics[trim=432 0 0 0, clip, width=0.245\linewidth]{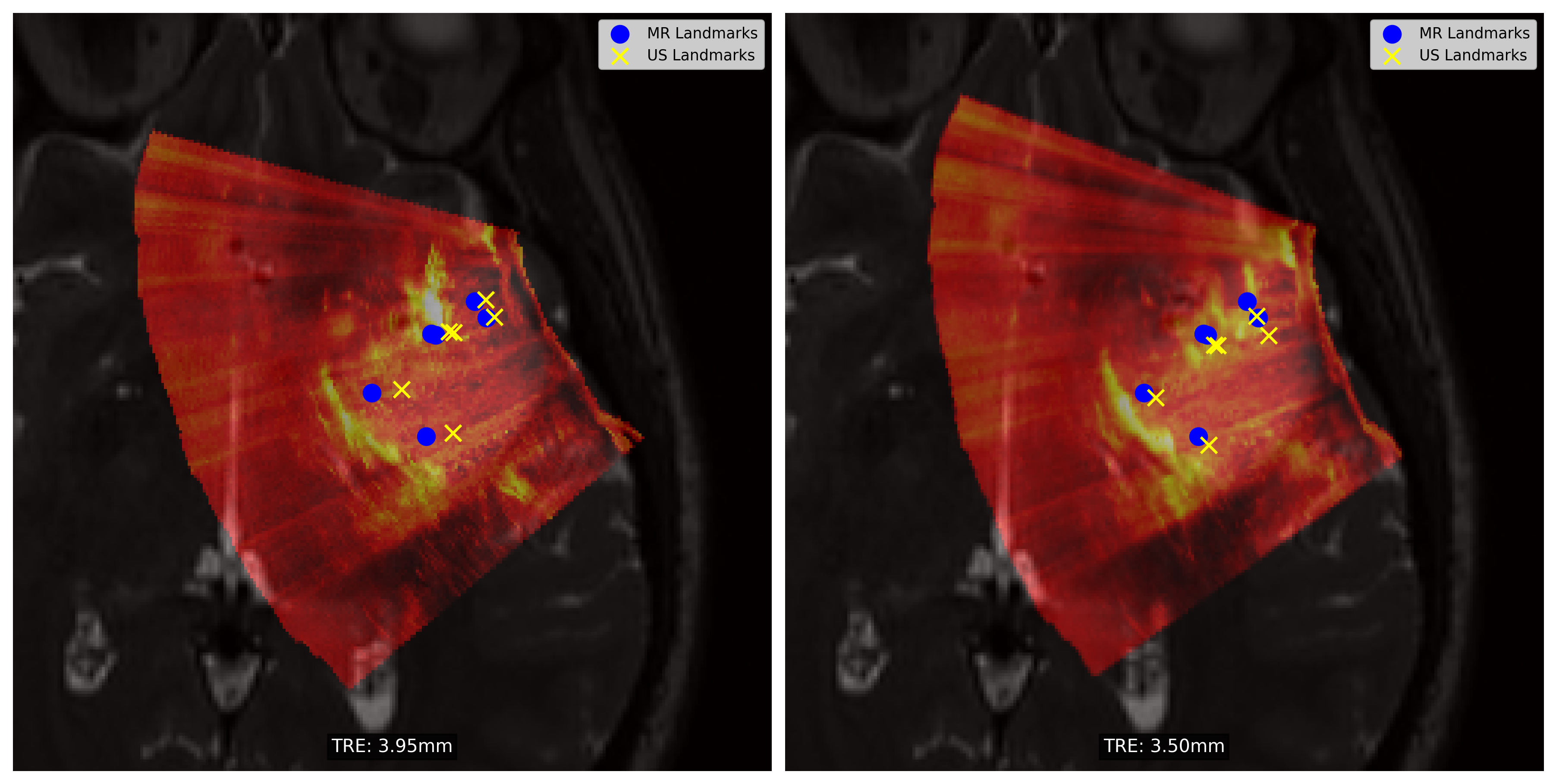}
    \includegraphics[trim=432 0 0 0, clip, width=0.245\linewidth]{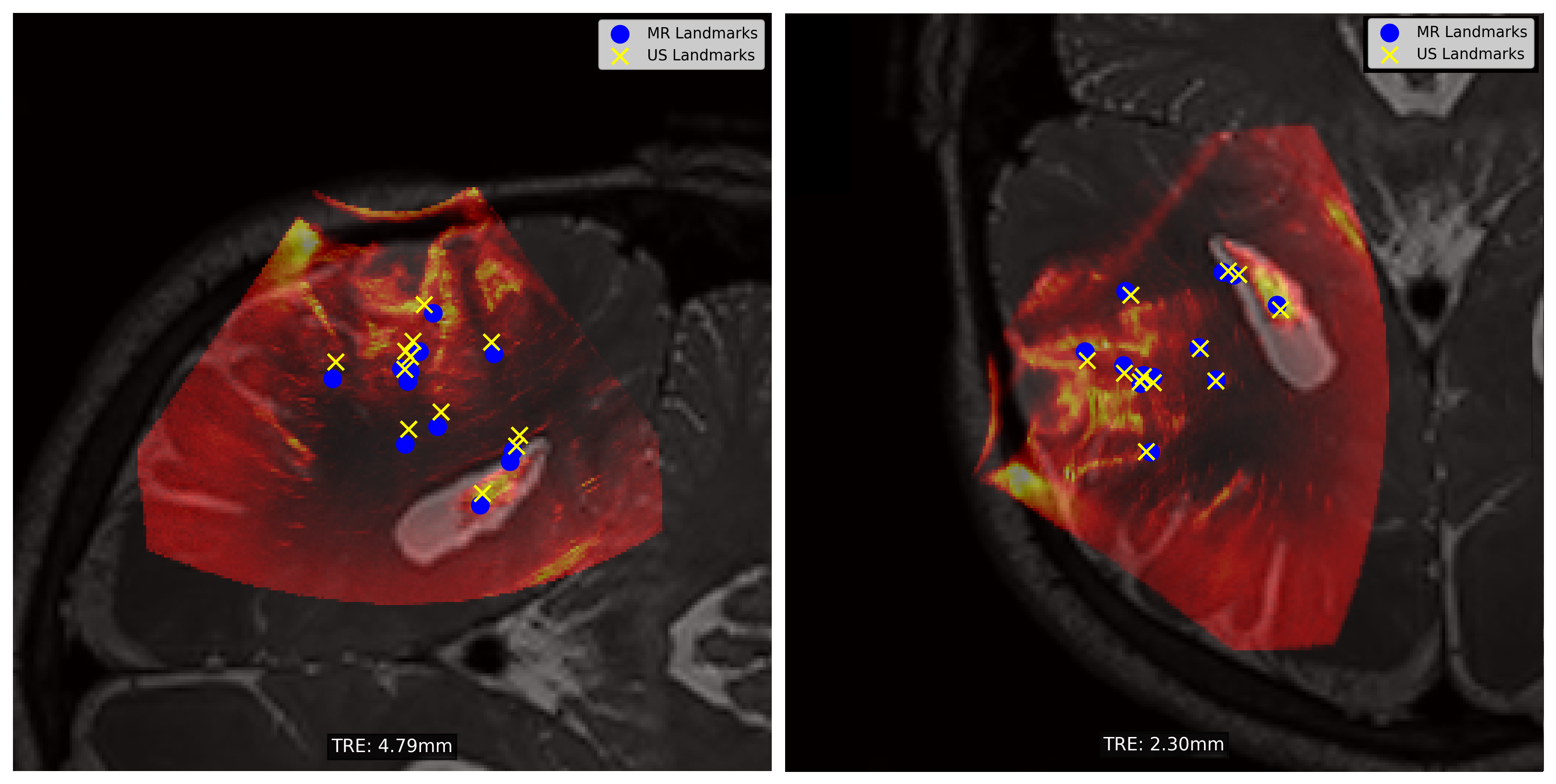}
    \includegraphics[trim=432 0 0 0, clip, width=0.245\linewidth]{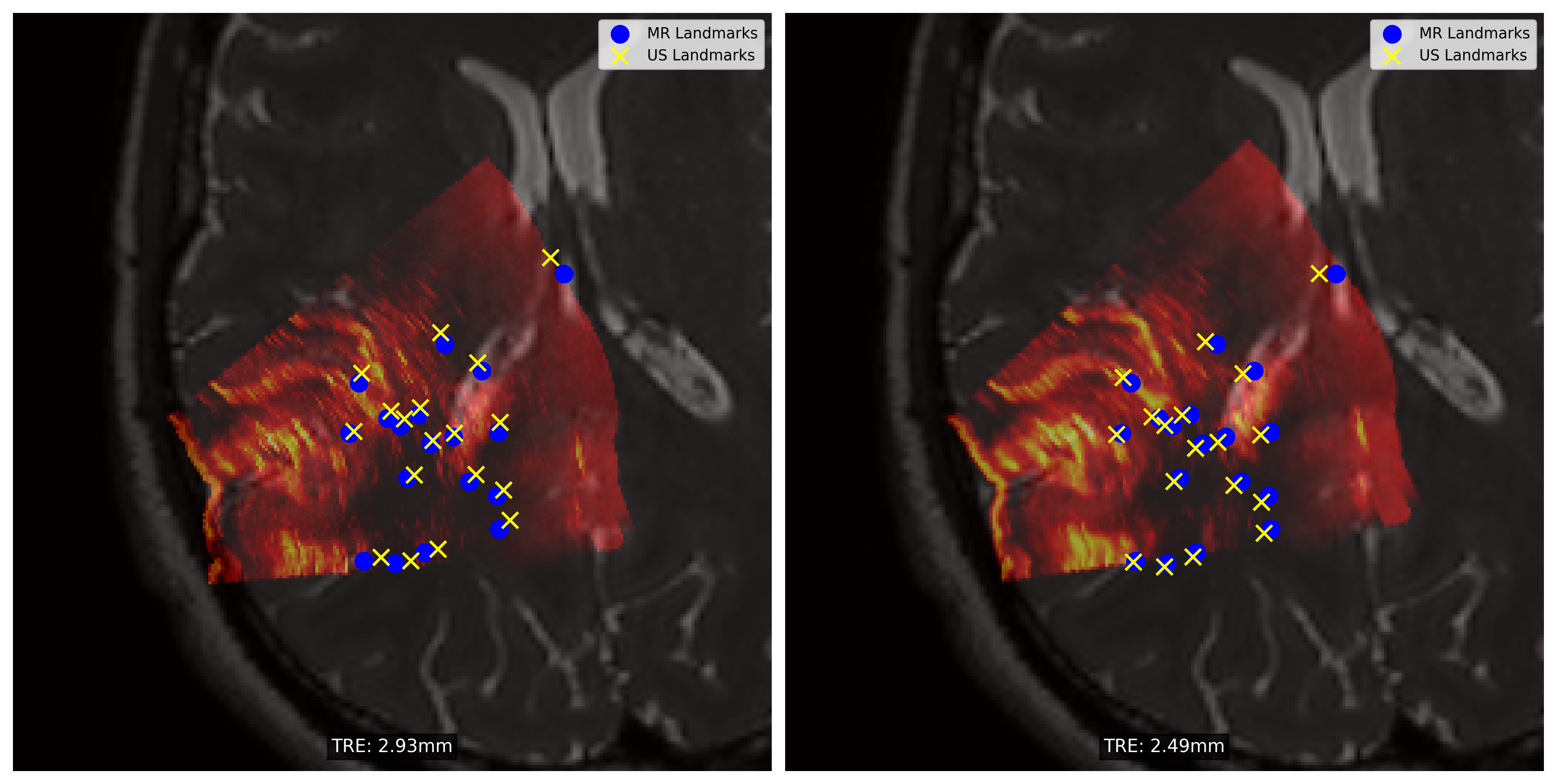}
    \includegraphics[trim=432 0 0 0, clip, width=0.245\linewidth]{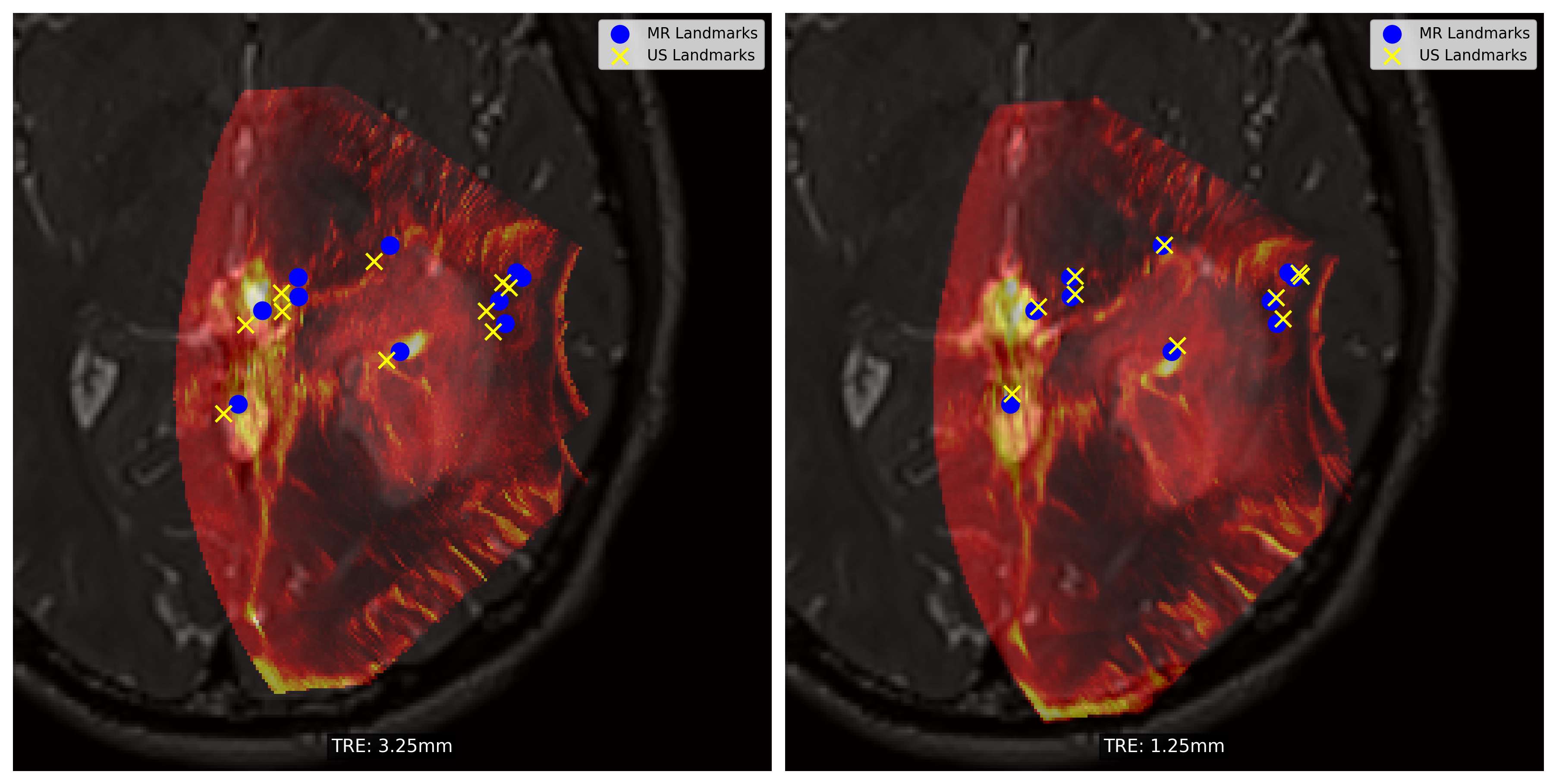}
\caption{Qualitative results of MR-iUS registration showing T2-weighted MR (grayscale) and the overlaid iUS (color wash, e.g., red), with MR landmarks (e.g., blue dots) and iUS landmarks (e.g., yellow crosses) indicating TREs. Our method shows good alignment of anatomical structures and small TREs between the landmarks.}
\label{fig:registration_visualization}
\end{figure*}

\subsubsection{Results}
The results for 4 cases from the validation dataset are presented in Table~\ref{tab:registration_comparison}, and ranked by their mean TREs (corresponding to the leaderboard rank).

\added{An interesting results is next-gen-nn that shows that MIND performs poorly when used as a standalone local descriptor for direct matching without spatial regularization or global optimization, particularly under large displacements. In contrast, its strong performance within variational registration frameworks reflects fundamentally different usage.}\rev{R1}

Our approach achieves a mean TRE of $2.385 \pm 0.397$~mm, placing third in the challenge leaderboard. Despite using only a rigid transformation model and relying exclusively on sparse keypoint correspondences, our method demonstrates competitive performance relative to several learning-based and deformable strategies. 
\added{We emphasize that rigid registration was chosen for robustness to outliers and large appearance changes in iUS. Extending the framework to non-rigid registration is an important direction for future work.}\rev{R3}

\begin{table}[htpb]
\centering
\caption[Results of registration methods on the ReMIND2Reg validation set]{Results of registration methods on the ReMIND2Reg validation set.}\label{tab:registration_comparison}
\begin{tabular}{r|l|c}
\toprule
\textbf{Rank} & \textbf{Method} & \textbf{Mean TRE (mm)} \\
\midrule
\rowcolor{LightGray}
1 & VROC & $1.903 \pm 0.582$ \\
2 & next-gen-nn & $1.969 \pm 0.459$ \\
\rowcolor{LightGray}
\textbf{*} & \textbf{Ours (Rigid, Keypoint-Based)} & $2.385 \pm 0.397$ \\
3 & Coarse-to-Fine w/ Style Transfer & $2.419 \pm 0.669$ \\
\rowcolor{LightGray}
4 & Topological Higher-Order MRF & $3.680 \pm 0.620$ \\
\bottomrule
\end{tabular}
\end{table}

To qualitatively assess the registration accuracy, we visualize landmark alignments after applying our method in Figure~\ref{fig:registration_visualization} with four representative cases showing overlaid MR and iUS volumes.
Our method achieves visually coherent registration even in challenging scenarios involving significant tissue deformation and modality-specific artifacts. The iUS FoV aligns well with the corresponding anatomical regions in the MRI, and the landmark correspondences show consistent spatial relationships. The registration quality is particularly evident in cases where complex anatomical structures, such as ventricular boundaries and tissue interfaces, maintain their expected spatial relationships after transformation.

The results demonstrate the potential of our keypoint-based registration framework, offering a robust and accurate solution without the need for complex optimization strategies or manual initialization, and providing clear visual cues for the registration process through matched keypoints.

\added{Figure~\ref{fig:USs} qualitatively compares US images before and after tumor resection, demonstrating the substantial anatomical and appearance changes in intraoperative US. Despite these changes, the proposed matching-by-synthesis strategy maintains accurate alignment.}\rev{R2,R3}

\section{Discussion, Limitations and Conclusion}
We presented a novel, fully 3D, rotation-invariant and cross-modal keypoint descriptor specifically
designed for the task of MRI–to–iUS volume matching. Our method bridges two highly distinct
modalities and goes beyond contrast-agnostic or monomodal approaches. To the best of our
knowledge, this is the first keypoint descriptor tailored to this domain. Our method consistently
outperforms existing state-of-the-art descriptors and matching techniques, and achieves
registration accuracy on par with leading methods from recent benchmark challenges. In addition
to accuracy, our approach offers key advantages in terms of usability and interpretability.
Matched and mismatched keypoints can be directly visualized, allowing clinicians to assess
anatomical plausibility and registration confidence. Furthermore, our registration pipeline does
not require manual initialization, which is commonly necessary in optimization-based methods
and presents a major barrier to clinical adoption.

\added{Beyond quantitative accuracy, our results highlight the importance of explicitly separating
the roles of matching and registration in cross-modal alignment. While registration methods can
often compensate for weak local similarity through global optimization and regularization, their
performance ultimately depends on the availability of reliable and well-distributed
correspondences. Our findings demonstrate that improving correspondence quality and spatial
coverage at the matching stage leads to more robust downstream registration, particularly under
large cross-modal appearance changes and post-resection anatomical variability.}\rev{R1,R3}

Our approach has a few limitations. 
\added{A fundamental design choice of this work is its patient-specific formulation. Consequently,
we do not evaluate inter-patient generalization for matching or registration, as the objective is
not to learn a population-level descriptor but rather to optimize correspondences for a given
patient using their own preoperative imaging. This choice is consistent with prior evidence
showing that patient-specific models can outperform patient-agnostic ones in deformable
registration and segmentation tasks, particularly in the presence of large anatomical and
appearance variability.}\rev{R3}

Second, the patient-specific training strategy, while improving accuracy, requires approximately
five hours of training. However, since preoperative MRI is routinely acquired well in advance of
surgery, this offline training time is clinically acceptable. Moreover, recent findings suggest
that given a pretrained model, rapid fine-tuning on a new patient’s pre-operative images can
retain high performance while reducing training time substantially~\cite{gopalakrishnan2025rapid}.

\begin{figure}[h!]
 \centering
  \subfloat{\includegraphics[clip, trim=20 20 20 20, width=0.495\linewidth]{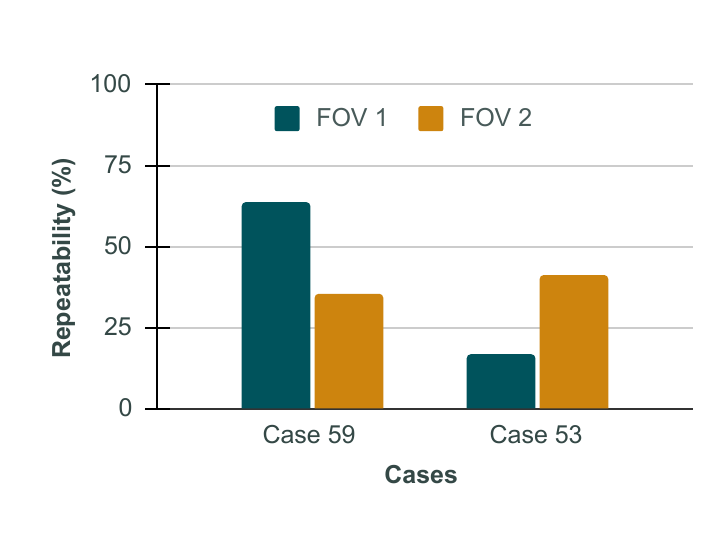}}
    \hfill
  \subfloat{\includegraphics[clip, trim=20 20 20 20, width=0.495\linewidth]{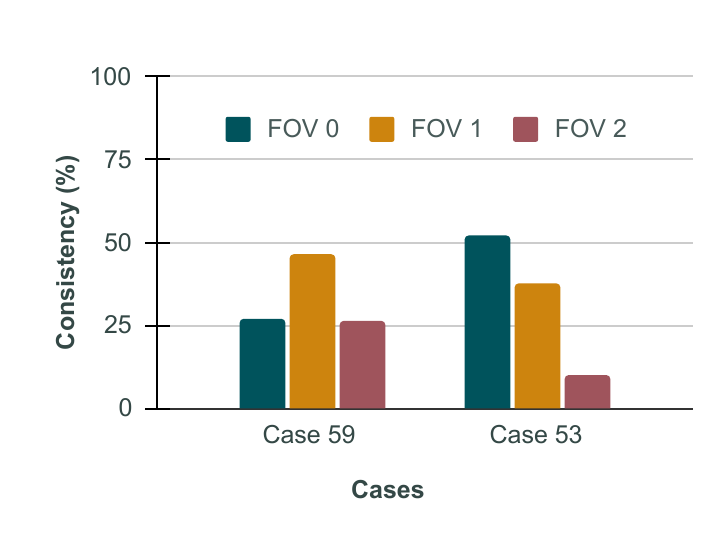}}\\
 \subfloat{\includegraphics[clip, trim=20 20 20 20, width=0.495\linewidth]{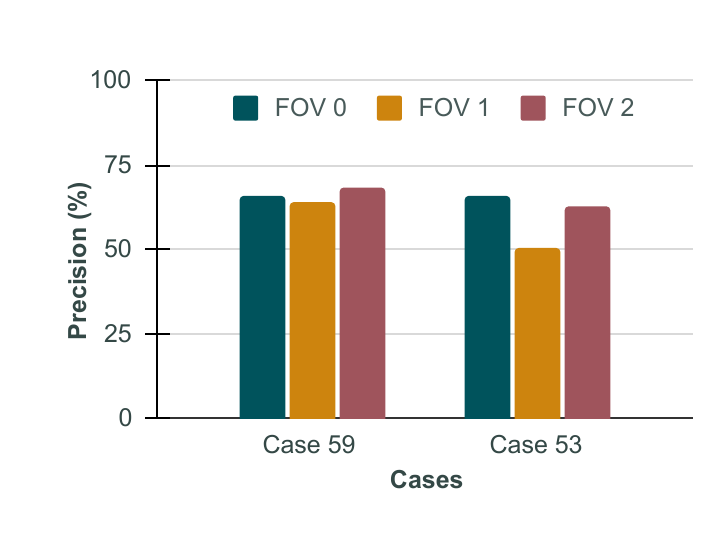}}
    \hfill
  \subfloat{\includegraphics[clip, trim=20 20 20 20, width=0.495\linewidth]{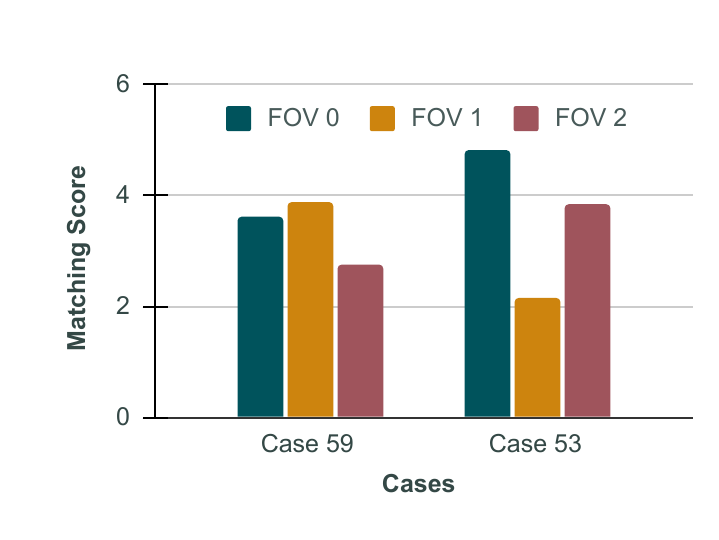}}
 \caption{Evaluation of descriptor robustness across varying iUS FoVs. Despite changes in FoV,
 keypoint matching remained consistent and repeated, with good precision and accuracy,
 indicating generalization beyond training conditions.}
 \label{fig:fov_invariance}
\end{figure}

Third, our synthesis-based training leverages a fixed FoV prior to generate synthetic US images.
This raises concerns about potential performance degradation when the iUS FoV during surgery
deviates from the simulated training FoV. To investigate this, we conducted an experiment using
two cases with three distinct iUS FoVs per case. As shown in
Figure~\ref{fig:fov_invariance}, the model demonstrates high repeatability and consistency across
varying FoVs, while maintaining precision and accuracy. Keypoints detected in the original FoV
were reliably recovered in the other FoVs, and a large fraction of matched keypoints were
consistent across all three views. This suggests that the learned descriptor generalizes well
beyond the synthetic training conditions and remains robust under practical variations in iUS
coverage.

\added{Importantly, robustness in this work is assessed through clinically realistic sources of variability, including substantial pre- and post-resection anatomical changes and variations in iUS FoVs, rather than through cross-dataset evaluation. Existing iUS datasets do not
provide paired preoperative MRI and pre-dural US required for this setting, and thus do not allow evaluation under the same conditions. We therefore view the presented experiments as a more clinically relevant assessment of robustness for the targeted application.}\rev{R2,R3}

Finally, while we tested our method using rigid registration on resected, non-rigid tissue, keypoint-based methods are inherently compatible with non-rigid registration pipelines. They can be integrated as sparse constraints within biomechanical models or B-spline
frameworks~\cite{Haouchine2022}. Moreover, they are well-suited to accommodate topological changes such as tumor resection, where matched keypoints can inform tissue stress estimation and guide MRI updates. Future work will explore these extensions, as well as downstream tasks including slice-to-volume registration for freehand ultrasound reconstruction and segmentation propagation through keypoint-guided interactive editing.

\section*{Acknowledgments}
This work was supported by the National Institutes of Health (R01EB032387, R01EB034223, and K25EB035166). R.D. received a Marie Skłodowska-Curie grant No 101154248 (project: SafeREG). The research leading to these results has received funding from the French government under management of Agence Nationale de la Recherche as part of the ``France 2030'' program (reference ANR-23-IACL-0008, PRAIRIE-PSAI) and as part of the "Investissements d'avenir" program (reference ANR-19-P3IA-0001, PRAIRIE 3IA Institute; and  reference ANR-10-IAIHU-06).

\bibliographystyle{IEEEtran}
\bibliography{references}

\end{document}